# ARTIFICIAL INTELLIGENCE-BASED PROCESS FOR METAL SCRAP SORTING

Maximilian Auer, Kai Oßwald, Raphael Volz, Jörg Woidasky

Pforzheim University, School of Engineering, 75175 Pforzheim, Germany
auermaxi@hs-pforzheim.de

## ABSTRACT

*Machine learning offers remarkable benefits for improving workplaces and working conditions amongst others in the recycling industry. Here e.g. hand-sorting of medium value scrap is labor intensive and requires experienced and skilled workers. On the one hand, they have to be highly concentrated for making proper readings and analyses of the material, but on the other hand, this work is monotonous. Therefore, a machine learning approach is proposed for a quick and reliable automated identification of alloys in the recycling industry, while the mere scrap handling is regarded to be left in the hands of the workers. To this end, a set of twelve tool and high-speed steels from the field were selected to be identified by their spectrum induced by electric arcs. For data acquisition, the optical emission spectrometer Thorlabs CCS 100 was used. Spectra have been post-processed to be fed into the supervised machine learning algorithm. The development of the machine learning software is conducted according to the steps of the VDI 2221 standard method. For programming Python 3 as well as the python-library sklearn were used. By systematic parameter variation, the appropriate machine learning algorithm was selected and validated. Subsequent validation steps showed that the automated identification process using a machine learning approach and the optical emission spectrometry is applicable, reaching a maximum F1 score of 96.9 %. This performance is as good as the performance of a highly trained worker using visual grinding spark identification. The tests were based on a self-generated set of 600 spectra per single alloy (7,200 spectra in total) which were produced using an industry workshop device.*

## 1. INTRODUCTION

The sorting of medium-value scrap is labor intensive and requires experienced and skilled workers. It is usually carried out at a hand-work place. This work is monotonous on the one hand; on the other hand, it requires high concentration while analyzing the results. Typical analytical methods in the field are X-ray fluorescence analysis, grinding spark testing and the spark and arc spectroscopy. X-ray fluorescence analysis is applied often, since it is easy to use, and readings from the hand-held devices are easy to read. However, this analysis may take up to 30 seconds, and sample pretreatment such as coating removal may be required, and EHS concerns may arise from X-ray use. Grinding spark testing, on the other hand, is easy and quick, but requires highly trained and experienced staff. Therefore, the aim of the research work was to develop a machine learning approach which automatically, quickly and reliably identifies tool and high-speed steel alloys. As a basis for analysis, the spectra of twelve different alloys from the field have been selected and analyzed, measuring their specific electromagnetic irradiation [1]. As each chemical element in the alloy emits a unique spectrum, by analyzing a sample's spectrum its single elements and their concentration may be identified [2]. The spectra were measured using emission spectrometry. The identification process was intended to be conducted by a machine learning algorithm after supervised learning [3].

The overall development goal was to substitute the current identification process using a machine learning algorithm along with the optical emission spectroscopy already applied in the field. The aim of this paper is to present the development of a machine learning algorithm to identify spectra, measured by a spectrometer. The algorithm must have an accuracy of more than 90 % for the individual alloys and an accuracy of more than 95 % for the cluster of an alloy, e.g. high-speed steel with or without cobalt. Furthermore, the time required for analysis must be similar to currently used techniques, ideally not exceeding 1 s.

## 2. STATE OF THE ART

Currently, there are various techniques used to sort metal scrap. They reach from hand-held solutions using X-ray fluorescence analysis to a visual identification by grinding spark testing or an optical spectrum. There is a new solution for nonferrous metals and alloys sorting them in a single pass, which is based on laser-induced breakdown spectroscopy [4]. In a broader range of sorting waste, there are some companies providing solutions based on Artificial Intelligence and optical sorting [5][6]. These solutions are able to differentiate between various materials like metals, wood, or plastic. Nevertheless, a sorting of specific alloys is not conducted by these machines.

## 3. MATERIALS AND METHODS

For data acquisition, a hand-workplace with a spectroscope (R. Fuess Berlin-Steglitz) was used. It was equipped with a spectrometer Thorlabs CCS 100 (350 nm to 700 nm measuring range). The samples were taken from the field, belonging to twelve different steel alloys, representing tool and high-speed steels. For each alloy, three physical samples with different shapes and sizes were selected for analysis. The alloys tested were M1, M2, T1, P9, M35, M36, M42, T4, T42, D2, H10 and H13. The samples were analyzed upfront using X-ray fluorescence analysis to properly identify the alloys.

Atomic emission of the metals was measured by the Thorlabs CCS 100 spectrometer and saved as a csv-file by the software Thorlabs OSA in relative intensity units (range 0 to 1). Spectrometer and arc generation configuration have been adjusted during the data acquisition process to acquire readable datasets. A good figure of usable spectra was found with measurements using an integration time of 100 ms and an angle of 1° between the fiber optic and the spectroscope.

The csv-files acquired by the Thorlabs spectrometer had to be processed to be feasible for supervised machine learning. Thus, all lines containing unnecessary information like time and date were removed. Moreover, the intensities of each spectrum were analyzed, and the spectrum was deleted if one of the following conditions was met: All intensities are smaller then 0.1; there is no wavelength with an intensity bigger than 0.2; or the spectrum contains at least one intensity reaching the value one. In a final step, all spectra were checked visually as a final data quality control. These datasets have been labelled with an index for the corresponding alloy, transposed, and eventually merged to one single csv-file, ready for the machine learning algorithm, creating miscellaneous sub-sets if needed.

For programming, the programming language Python 3 and the open-source Python libraries Sklearn, Pandas, NumPy and matplotlib were used.

## 4. DEVELOPMENT

As a first step of the conceptual design, the main function of the process was defined to "identify an alloy by using a machine learning algorithm". This function was sub-divided into three sub-functions "acquire data", "preprocess data" and "train the machine learning algorithm", as shown in Figure 1. The sub-functions required for the function "train the machine learning algorithm" are "load data", "split data", "scale data", "train algorithm", "test algorithm", "make predictions" and "export model".

| function tree for "identify an alloy by using a machine learning algorithm" | | |
|---|---|---|
| main function | function | sub-function |
| identify an alloy by using a machine learning algorithm | acquire data | place sample<br>light electrical arc<br>start measurement<br>end measurement |
| | preprocess data | clean-up data<br>label data<br>format data |
| | train the machine learning algorithm | load data<br>split data<br>scale data<br>train algorithm<br>test algorithm<br>make prediction<br>export model |

Figure 1. Function tree for the main function "identify an alloy by using a machine learning algorithm"

The principal solution for "data acquisition" was already defined by the selection of the measurement device. "Data acquisition and preprocessing" have been presented in chapter 2 already. For "training of the machine learning algorithm", in the subfunctions "load data" and "split data" commands from the python-libraries pandas and sklearn were used. For splitting the data, either a manual train test split is conducted, or the command for the k-fold cross-validation is used. For "scale data", various scalers included in the sklearn-library were tested, including the StandardScaler, MinMaxScaler, MaxAbsScaler, RobustScaler, Normalizer, or no scaler.

The given problem of identifying the alloy is a classification problem, so algorithms chosen were the k-nearest-neighbor, logistic-regression, multi-layer perceptron, decision tree, random forest, SVM (support vector machine) with rbf-kernel and linear SVM. For testing and making predictions commands were used which are included in the sklearn-libraries. The final model was exported using sklearn-command as well. For identifying the best suitable algorithm, a practical testing approach is chosen, applying supervised learning [7].

## 5. APPLICATION AND RESULTS

In a first approach, all algorithms and scalers pre-selected in chapter 3 were tested. For each alloy 400 spectra were used, using in total 4800 datasets. The mean $F_1$ score was used to compare the performance of each classifier and scaler. For the train-test-split, the 10-fold cross-validation was used.

The results of the test are presented in Figure 2, showing the mean $F_1$ score for each algorithm and scaler. Furthermore, the standard deviation is given in this graph [8]. Comparing the effects of the scalers shows some algorithms sensitivity to the scalers used: Algorithms like k-nearest neighbor, decision tree and random forest require the use of the Normalizer. Other algorithms like the SVM with rbf-kernel perform best using a scaler centering the data, like StandardScaler or RobustScaler.

To select the algorithm with the best performance, the $F_1$ scores were compared. Due to the fact, that the decisions tree ensemble method, the random forest, performs better, this classifier was not considered any longer. Also, the SVM with a rbf-kernel was not considered any longer since the parameter C with a value of 30 is rather high. The good performance with a $F_1$ score of

98.8 % may be attributed to overfitting of the machine learning algorithm. Moreover, the linear SVM, based on a linear kernel, has a 0.3 percent point higher $F_1$ score of 99.1 %.

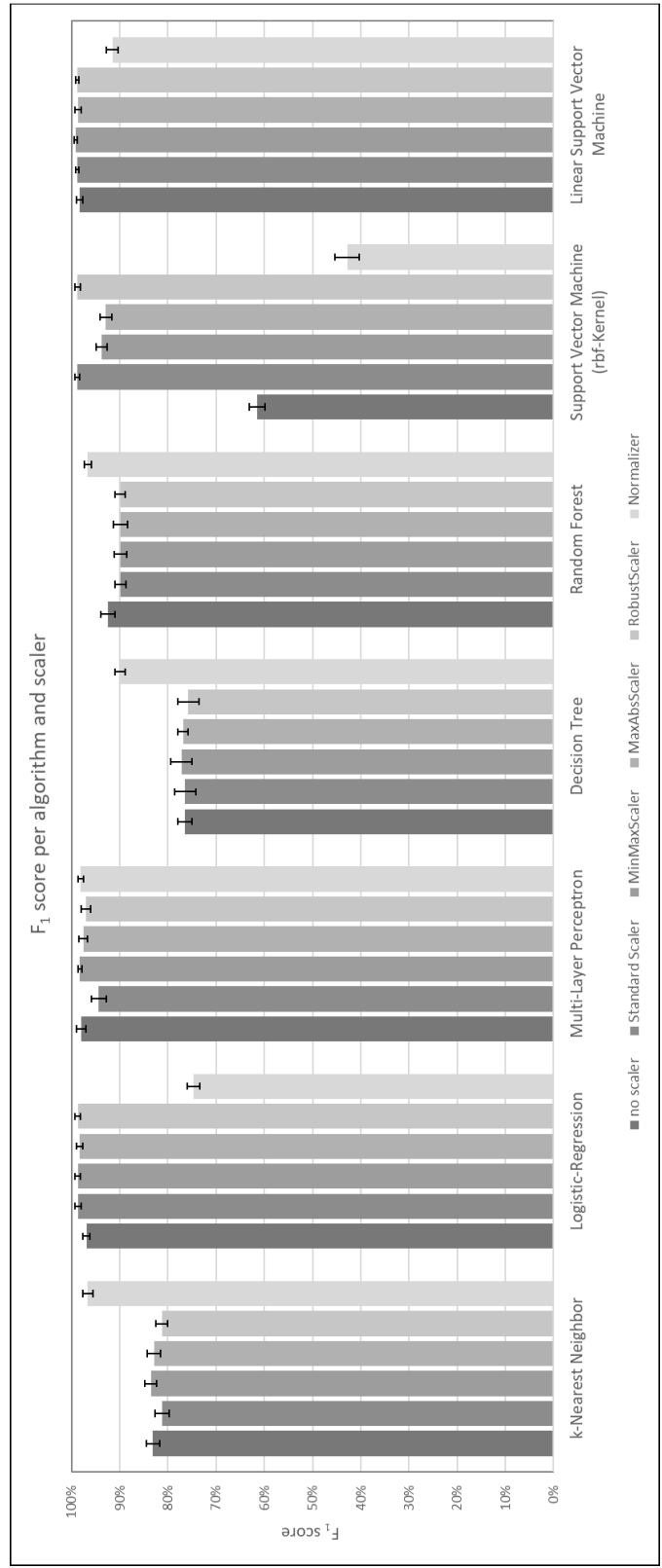

Figure 2. Comparison of the $F_1$ score per algorithm and scaler

Another factor considered in algorithm selection is the time consumed to make a prediction. Therefore, the time of testing and making prediction is measured. In this category, the k-nearest neighbor performed poorly, taking 22.8 s for all 1200 test spectra. All other algorithms needed less than 1 s. For further work, the three best performing machine learning algorithms were chosen: Logistic regression (98.9 %), multi-layer perceptron (98.3 %) and the linear SVM (99.1 %).

In the second test these three algorithms were used to make predictions for an unknown sample as well as observing the influence of a growing number of training data. Therefore, two physical samples of each alloy were used for training and one sample was used for testing. The number of training data varied between 5 and 450 spectra per alloy, whereas for testing always 150 spectra per alloy were used.

The results given in Figure 3 show that a growing number of training data has a positive effect on the performance of the algorithm, independently from the machine learning algorithm itself. With already 50 spectra for each alloy as training data, resulting in a total of 600 spectra, a $F_1$ score from up to 92.0 % is achieved using the linear SVM. Doubling the training data results in a 2.3 percentage point increase. The maximum achieved $F_1$ score is 96.6%. Comparing the $F_1$ scores of each algorithm, the linear SVM achieves the best performance ($F_1$ score = 96.9 %), followed by the logistic regression classifier (96.0 %) and the multi-layer perceptron (93.3 %). Therefore, further tests were conducted using the linear SVM.

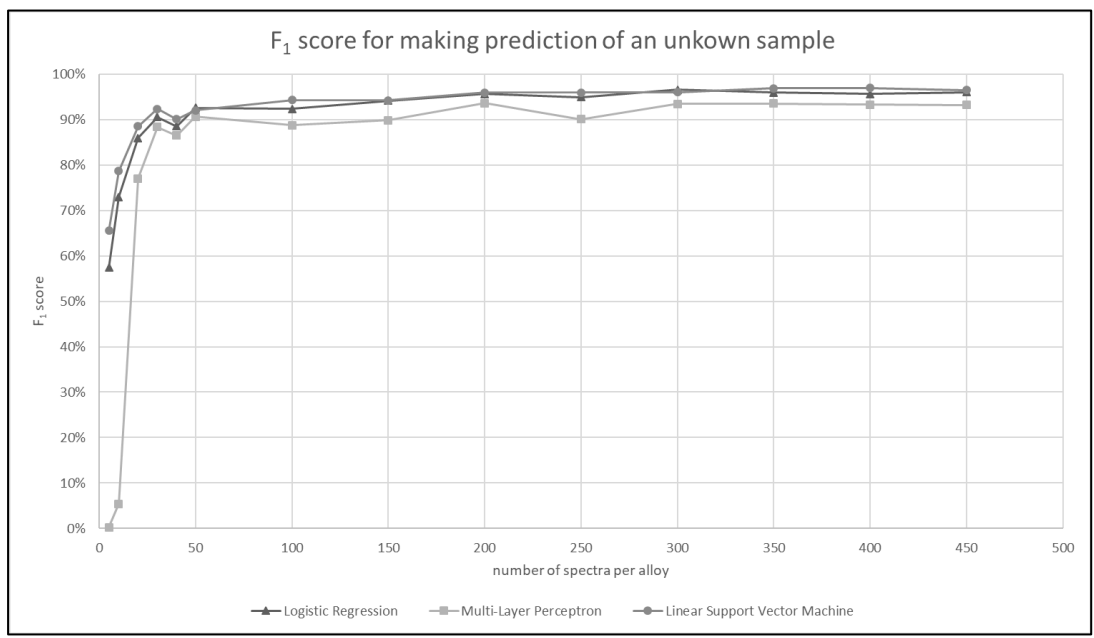

Figure 3. $F_1$ score for making predictions of an unknown sample

In a variation of the test, the $F_1$ score was evaluated considering predictions which were made for only one alloy. In case of the linear SVM this means, the algorithm identifies only one model with a positive distance to the hyperplane. For training, 400 spectra per alloy were used. From the tested spectra, 88.4 % were clearly assigned to one alloy. From those clearly assigned predictions, 98.3 % were correct.

In the next test, the parameters of the linear SVM were optimized to obtain a better performance. Therefore, the data was split into training data, containing the 400 spectra of two samples per alloy, and test data, containing the remaining samples with 150 spectra each. First, the performance of the 10-fold cross validation was calculated. The best performing model was selected to make prediction for the test data. For optimizing the performance, two different pa-rameters were adjusted: the loss function and the parameter C. The loss functions tested were hinge and squared-hinge, the values for C were 0.1, 1.0, 10, 100 and 1000. The parameters with the best performance of the test data were selected for further use.

The linear SVM performed best with the loss function squared_hinge and the value 1 for C. In general, it was observed, that the squared_hinge function performed always slightly better than the hinge function. For a higher value of C, the $F_1$ score of the 10-fold cross validation increased, whereas the $F_1$ score of the test data decreased. The higher the value of C is, the more the linear SVM tends to overfit [8].

For the last test the optimized linear SVM was validated. Therefore, test data from two new physical samples, 100 per sample and per alloy, were acquired. For training, the 400 spectra per alloy of the previous test were used. The performance of identifying an individual alloy as well as the cluster of an alloy were measured, using the $F_1$ score. The three clusters were tool steels, high-speed steels with cobalt, and high-speed steels without cobalt.

The $F_1$ score for identifying an individual alloy only reached 49.1 %, a remarkably low value compared to the result of the previously conducted test 2. For analysis, a confusion matrix was generated (Table 1). The confusion matrix shows, that alloys are often confused with an alloy of the same cluster. Analyzing the samples a second time using the X-ray fluorescence analysis and comparing the composition showed that the expected composition limits of the single samples were often exceeded. These off-spec samples were: M2 (no. 104), T1 (no. 105, 106), P9 (no. 107, 108), T42 (no. 113, 114), T4 (no. 115, 116). In total, only ten samples were identified explicitly. Therefore, this poor result is not representative and cannot be used for a validation of the algorithm. Nevertheless, analyzing the cluster, a $F_1$ score of 94.5 % was achieved. Only few alloys were misclassified to another cluster.

Table 1. confusion matrix for the identification of an individual alloy

| | | | predicted alloy | | | | | | | | | | | |
|---|---|---|---|---|---|---|---|---|---|---|---|---|---|---|
| | | | HSS w/o Co | | | | HSS w/ Co | | | | | CrMoV | | |
| | | alloy | M1 | M2 | T1 | P9 | M35 | M36 | M42 | T4 | T42 | D2 | H10 | H13 |
| | | sample-no. | 101 \| 102 | 103 \| 104 | 105 \| 106 | 107 \| 108 | 109 \| 110 | 111 \| 112 | 113 \| 114 | 115 \| 116 | 117 \| 118 | 119 \| 120 | 121 \| 122 | 123 \| 124 |
| true alloy | HSS w/o Co | M1 | 101 \| 102 | 0\|0 | 0\|0 | 0\|0 | 0\|0 | 0\|0 | 0\|0 | 0\|0 | 0\|0 | 0\|0 | 0\|0 | 0\|0 | 0\|0 |
| | | M2 | 103 \| 104 | 0\|0 | 96\|10 | 0\|0 | 0\|0 | 0\|*37* | 0\|*16* | 0\|0 | 0\|*32* | 0\|0 | 0\|0 | 4\|5 | 0\|0 |
| | | T1 | 105 \| 106 | 0\|0 | *47*\|4 | 26\|1 | *16*\|*88* | 2\|0 | 0\|0 | 0\|0 | 9\|7 | 0\|0 | 0\|0 | 0\|0 | 0\|0 |
| | | P9 | 107 \| 108 | 4\|0 | 0\|0 | 0\|4 | 86\|93 | 5\|0 | 0\|0 | 1\|0 | 1\|0 | 0\|0 | 0\|0 | 0\|0 | 3\|3 |
| | HSS w/ Co | M35 | 109 \| 110 | 0\|0 | 1\|4 | 0\|0 | 0\|1 | 3\|0 | *18*\|*65* | *58*\|*27* | *20*\|2 | 0\|1 | 0\|0 | 0\|0 | 0\|0 |
| | | M36 | 111 \| 112 | 0\|0 | 0\|1 | 0\|0 | 0\|1 | 1\|0 | 96\|81 | 2\|4 | 1\|13 | 0\|0 | 0\|0 | 0\|0 | 0\|0 |
| | | M42 | 113 \| 114 | 0\|2 | 0\|0 | 0\|0 | 0\|0 | 0\|0 | 0\|0 | 100\|96 | 0\|2 | 0\|0 | 0\|0 | 0\|0 | 0\|0 |
| | | T4 | 115 \| 116 | 0\|0 | 0\|0 | 1\|5 | 1\|4 | 4\|7 | 0\|0 | 0\|0 | 49\|64 | *43*\|*20* | 0\|0 | 0\|0 | 2\|0 |
| | | T42 | 117 \| 118 | 0\|0 | 0\|0 | 0\|0 | 3\|2 | 0\|0 | *46*\|*58* | 0\|0 | 0\|0 | 51\|40 | 0\|0 | 0\|0 | 0\|0 |
| | CrMoV | D2 | 119 \| 120 | 0\|0 | 0\|0 | 0\|0 | 0\|0 | 0\|1 | 0\|0 | 0\|0 | 0\|3 | 0\|0 | 100\|96 | 0\|0 | 0\|0 |
| | | H10 | 121 \| 122 | 0\|0 | 0\|0 | 0\|0 | *15*\|0 | 0\|0 | 1\|0 | 3\|0 | 0\|0 | 0\|0 | *15*\|0 | 66\|0 | 0\|0 |
| | | H13 | 123 \| 124 | 0\|3 | 0\|0 | 0\|0 | 0\|0 | 0\|0 | 0\|0 | 0\|0 | 0\|0 | 0\|0 | *93*\|*87* | 5\|0 | 2\|10 |

In the implementation phase, the functional-logical design was transformed into a software solution with the trained machine learning algorithm.

## 6. CONCLUSION

The aim of the research work was to develop a machine learning approach for the alloy identification in scrap sorting. Twelve tool and high-speed steels from the field were selected to be identified by their spectra induced by electric arc. For data acquisition the optical emission spectrometry was used. The spectra were fed to the machine learning algorithm. In various tests, the appropriate algorithm was selected and validated.

The tests were conducted using the spectra of three samples per alloy. The measured relative intensity of one sample varies often, although the underlying sample composition was identical. This limited data variation may cause an overfitting of the machine learning algorithm. This effect was actually observed in the first two tests: Test 1 was used to select algorithms which were applicable for the identification problem in general. In this test, for training and testing spectra from all samples were analyzed using the 10-fold cross validation. The good performance of up to 99.1 %, using the linear SVM, was probably caused by the limited data variation. Thus, the split of training and test data was modified in test 2: Two samples of each alloy were used for the training, the third sample available was used for testing. Thereby, the performance value dropped to 96.9 % for the linear SVM. This performance drop confirmed the previous hypothesis of a possible overfitting. The second test showed as well that with a growing number of training data, the performance of the algorithm improved.

For a validation of these tests, test 4 was conducted. For this test, for each alloy two new samples were selected and 100 spectra from each sample were acquired and tested using the linear SVM trained with the training data from test 1. This algorithm now was able to identify only 49.1 % of these spectra correctly. This probably was caused by the actual physical alloy composition variation, exceeding the standard composition limits of the alloys. This effect makes it difficult to assign them clearly to one type of alloy. However, assignability to alloy clusters (tool steel/HSS steel with cobalt/HSS steel without cobalt) worked well (94.5 %). This test revealed the challenges of a sufficient learning database size, and of physical sample selection and preparation, as composition limits of the samples have to be observed.

The results of the research show that steel alloy identification using a machine learning approach and the optical emission spectroscopy is successful. Around 95 % of the spectra were identified correctly. So, the performance of a worker identifying the alloys using visual grinding spark identification was reached. Further steps will extend the database with additional alloys and samples. For the implementation in the operational business, the interfaces to the spectrometer and the output of the results have to be modified.

**Authors**

Maximilian Auer is studying Engineering and Management in master's at Pforzheim University. In August 2018, he finished his B. Sc. in Business Administrations and Engineering.

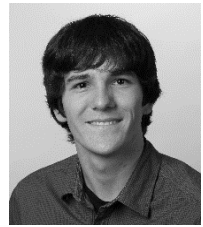

Kai Oßwald studied mechanical engineering at RWTH Aachen University with a major in production engineering. For his Ph. D. thesis, he developed laser drilling processes for diesel injection nozzles. He worked in different positions for Bosch in Stuttgart between 2001 and 2011. Since then he has been a professor for manufacturing processes at Pforzheim University.

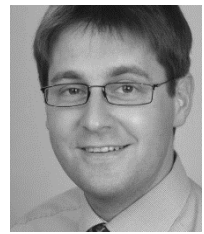

Raphael Volz studied informatics at Karlsruhe Institute of Technology (KIT). He obtained his doctorate summa cum laude at the KIT in 2004 researching Web Ontology Reasoning with Logic Databases. In 2009, he founded nogago GmbH. He also serves as chairman of the board for Volz Innovation GmbH. Since 2011, he is Professor for applied informatics at Pforzheim University.

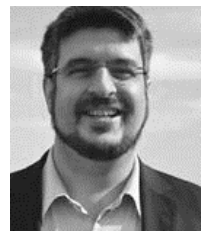

Joerg Woidasky graduated in Environmental Engineering from Berlin Technical University. He received his Ph. D. in mechanical engineering from the Stuttgart University in 2006 and worked more than two decades for the Fraunhofer Organisation for applied research. Since 2012, he is Professor for Sustainable Product Development at Pforzheim University.

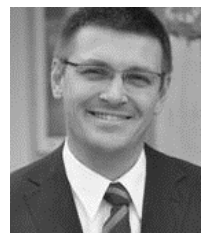